\documentclass{article}

\usepackage{arxiv}

\usepackage[utf8]{inputenc} 
\usepackage[T1]{fontenc}    
\usepackage{url}            
\usepackage{amsfonts}       
\usepackage{microtype}      
\usepackage{graphicx}
\usepackage{subcaption}
\usepackage{doi}

\usepackage{amsmath,accents}

\usepackage{cite}

\title{Gradient-based inverse lithography for EUV masks\\ via the waveguide method and\\ a physics-informed neural operator}

\date{}

\author{ \href{https://orcid.org/0000-0002-4930-1846}{\includegraphics[scale=0.06]{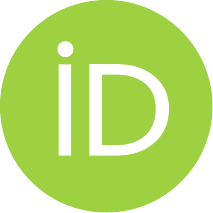}\hspace{1mm}Vasiliy A. Es'kin}\thanks{Corresponding author: Vasiliy Alekseevich Es’kin (\href{mailto:vasiliy.eskin@gmail.com}{vasiliy.eskin@gmail.com})} \\
	Department of Radiophysics, University of Nizhny Novgorod\\
	23 Gagarin Ave., Nizhny Novgorod 603022, Russia\\
	\href{mailto:vasiliy.eskin@gmail.com}{\texttt{vasiliy.eskin@gmail.com}} \\
	\And
	{Egor V. Ivanov} \\
	Department of Radiophysics, University of Nizhny Novgorod\\
	23 Gagarin Ave., Nizhny Novgorod 603022, Russia\\	\texttt{iev90078@gmail.com}
}

\renewcommand{\headeright}{}
\renewcommand{\undertitle}{}
\renewcommand{\shorttitle}{A preprint}

\renewcommand{\vec}{\bf}
\newcommand{\eps}{\varepsilon}

\begin{document}

\maketitle

\index{Es'kin, V.A.}
\index{Ivanov, E.V.}

\begin{abstract}	
Gradient-based inverse lithography technology~(ILT) for extreme ultraviolet~(EUV) masks is presented. A novel framework treats the differentiable waveguide method and the recently proposed waveguide neural operator~(WGNO) as end-to-end physics engines, recovering the permittivity of the absorber of the mask through automatic differentiation of the full forward diffraction model. Numerical experiments on realistic 2D and 3D absorbers of the mask (TaBN, La, U) at $\lambda{=}11.2$~nm show that the considered ILT methods make it possible to obtain a mask structure that achieves the desired field on the wafer.
\end{abstract}

\section{Introduction}

The relentless advancement of semiconductor technology, as described by Moore's law, has driven the feature sizes of integrated circuits into the angstrom regime. EUV lithography, utilizing wavelengths of $13.5$~nm and, as a future target, $11.2$~nm, is the cornerstone technology for manufacturing the most advanced semiconductor devices~\cite{it7}. Unlike deep ultraviolet systems, EUV systems rely on reflective photomasks, and the interplay between the patterned layer topography and obliquely incident electromagnetic waves causes mask three-dimensional effects, making the mitigation of these diffraction phenomena paramount for preserving pattern fidelity on the silicon wafer.

In processor manufacturing, optical proximity correction~(OPC) is applied to counteract the anticipated diffraction effects, and the most rigorous realization of OPC is inverse lithography technology~(ILT)~\cite{Pang2021,it1}. ILT formulates the optimization of the mask as an inverse problem, where the aim is to find the ideal mask topology that creates a desired intensity distribution on the wafer. While pixel-based ILT delivers superior pattern fidelity, it is burdened by the repeated invocation of a full-wave electromagnetic simulation, whose high computational cost limits its practical applicability.

Recently, deep neural networks, including supervised surrogates and physics-informed neural networks~(PINNs)~\cite{raissi2019physics,eskin2024causal}, have been proposed to accelerate the forward diffraction simulation~\cite{it1,medvedev2024}. However, supervised approaches require large pre-computed datasets and frequently fail to generalize across realistic mask geometries. To overcome these limitations, we have recently proposed a hybrid waveguide neural operator~(WGNO) that retains the physics-based pipeline of the rigorous waveguide method and replaces only the most computationally expensive step, namely the solution of the resulting linear system, by a multilayer perceptron~\cite{Eskin2025}.

In this paper, we extend the WGNO, and the underlying differentiable waveguide method, to the inverse lithography problem. We present an end-to-end gradient-based ILT framework in which the dielectric permittivity is optimized directly through the forward diffraction model by automatic differentiation, using either a pixel-wise density reparameterization or a Fourier-parameterized projection, both followed by hard binarization and validation by rigorous forward diffraction calculation. The same pipeline is combined with the WGNO, yielding a hybrid optimization scheme.

\section{Problem Formulation}

\begin{figure}[ht!]\centering
	\includegraphics[width=.43\textwidth]{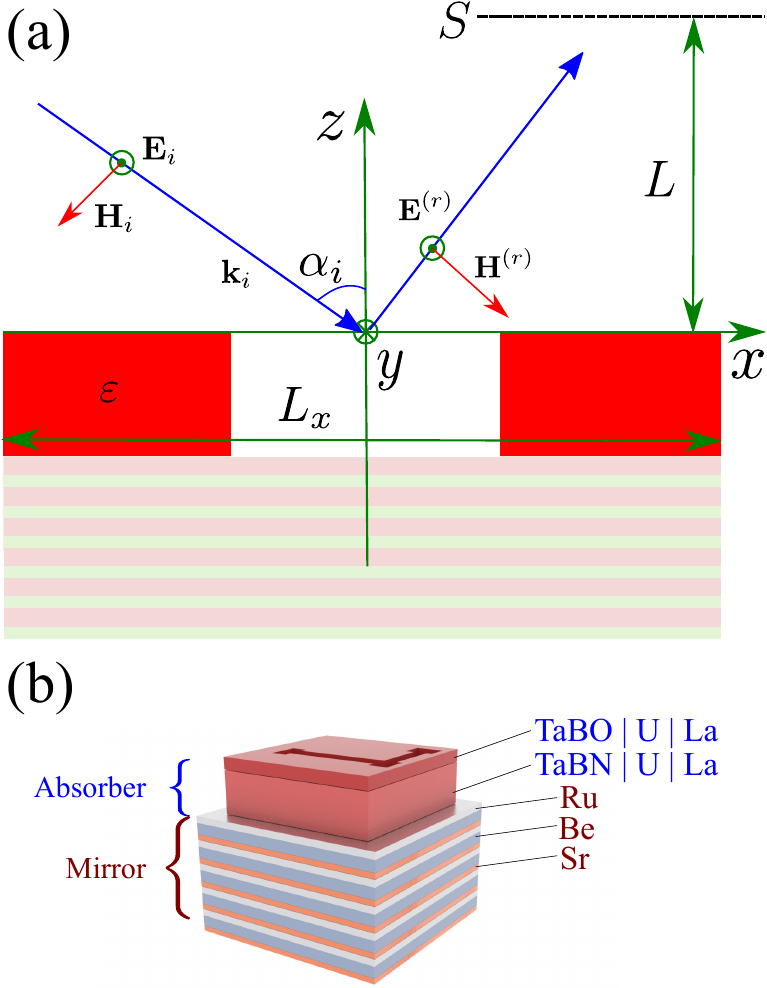}
	\caption{Geometry of the problem in the cross section $y {=} 0$ (a) and the stereometric view (b).}\label{fig1}
\end{figure}

Our consideration focuses on the two-dimensional formulation of the EUV mask diffraction problem that captures elongated circuit elements such as power delivery rails, interconnects, and clock distribution lines. The mask occupies an interval $[-D,0]$ along the $z$-axis of a Cartesian coordinate system $(x,y,z)$, as shown in Fig.~\ref{fig1}, and is periodic in the $x$ direction with period $L_x$. Each $j$th layer is filled with a nonmagnetic medium of permittivity $\eps_j(x)$ uniform in the $z$ direction. A TE-polarized electromagnetic monochromatic plane wave with angular frequency $\omega$ is incident on the mask at an oblique angle. The electric fields in the incident wave are given, with $\exp(i\omega t)$ time dependence dropped, by ${\vec E}^{(i)}={\vec y}_0 E_0\exp[-{\rm{i}}(k_{0;x}x-k_{0;z}z)]$, where $E_0$ is the electric field amplitude, $k_{0;x}$ and $k_{0;z}$ are the components of the wave vector ${\vec k}_0$ in free space ($k_0 = \left(k_{0;x}^2 + k_{0;z}^2\right)^{1/2}$, $k_0 = \omega /c$, where $c$ is the speed of light in free space), and the superscript $(i)$ denotes the incident wave. We denote by ${\vec E}^{(r)}$ the reflected electric field on a remote surface $S$, which is parallel to the $xOy$ plane and located at $z=L$ (in our numerical experiments we take $L=1$~m). The prescribed desired electric field and its intensity on the same surface are denoted as ${\vec E}^{(d)}$ and $I^{(d)}$ ($I^{(d)} = \left|{\vec E}^{(d)}\right|^2 $), respectively. The aim of the inverse lithography problem is to find the permittivity distribution of the layers $\eps_j(x)$ such that the intensity of the desired field coincides with the intensity of the scattered field on the target surface $S$:
\begin{align}
	\left|{\vec E}^{(r)}\right|^2 = \left|{\vec E}^{(d)}\right|^2.
\end{align}
We consider a mask consisting of a single absorber layer, the dielectric permittivity of which is denoted as $\eps(x)$, and a set of mask layers homogeneous along the $x$ axis.

The forward (diffraction) problem is governed by the time-harmonic Maxwell equations. Within each $j$th layer the magnetic field component $H_x$ satisfies the Helmholtz equation
\begin{equation}
	\Delta H_x + k_0^2\eps_j H_x = 0,
	\label{eq:helmholtz}
\end{equation}
supplemented by the periodicity condition ${\vec H}(-L_x/2,z)={\vec H}(L_x/2,z)$ and by the Sommerfeld radiation
condition on the top and bottom boundaries of the computational domain. The coupling of the fields between layers is enforced through the continuity of the tangential field components $H_x$ and $E_y$ at each lower ($z=z_{\min;j}$) and upper ($z=z_{\max;j}$) boundaries of the $j$th layer.

The inverse problem is to minimize the loss functional that measures the discrepancy between the intensities of the reflected and the desired fields on the remote surface. We formulate this optimization in two parametrizations. In coordinate space the problem is
\begin{align}\label{eq:opt_coord}
	& \eps(x)={\arg}\min_{\eps}\,\mathcal{L}(\eps),\\
	&\mathcal{L}(\eps)=\left.\!\left(\left|{\vec E}^{(r)}\right|^2 - \left|{\vec E}^{(d)}\right|^2\right)^2\!\right|_S.\notag
\end{align}
In the Fourier space the problem is the optimization of the truncated spectrum $\{\eps_m\}_{m=-N}^{N}$ of the permittivity or the Fourier-parameterized projection of permittivity
\begin{align}\label{eq:opt_fourier}
	&\{\eps_m\}={\arg}\min_{\{\eps_m\}}\,\mathcal{L}(\eps).
\end{align}
In both formulations the most computationally expensive operation is the calculation of the scattered field ${\vec E}^{(r)}$ for a given trial permittivity $\eps(x)$. The waveguide method provides an efficient discretization of this operator through an eigenvalue problem for each layer, followed by the solution of a global linear system~\cite{lucas1996efficient}. The central requirement for gradient-based ILT is that the entire forward operator be differentiable with respect to the design variables. We emphasize that this differentiability is the only property that distinguishes the present gradient-based ILT framework from conventional gradient-free or supervised-learning alternatives.

\section{Solution Methods}

\subsection{Differentiable Waveguide Method}

According to the waveguide method, the Helmholtz equation (\ref{eq:helmholtz}) is transformed into an algebraic eigenvalue problem by separating the transverse and longitudinal dependences, $H_x(x,z)=X(x)Z(z)$, and expanding $X(x)=\sum_{m=-N}^{N}B_m\psi_m$, where $\psi_m=\exp(-\mathrm{i}\kappa_x m x)$ and $\kappa_x=2\pi/L_x$. Substitution into (\ref{eq:helmholtz}) yields, after truncation to the band $|m|\le N$, the generalized eigenvalue problem
\begin{equation}
	\hat D{\vec B}=k_z^2{\vec B},\qquad
	\hat D_{nm}=k_0^2\eps_{n-m}-(\kappa_x n)^2\delta_{nm},
	\label{eq:eig}
\end{equation}
where $(2N{+}1)$ eigenvalues $k_{z;p}^2$ and eigenvectors ${\vec B}_p$ define the waveguide modes in every $j$th layer. The total field in the layer is a superposition of fields of the forward and backward modes. By satisfying the continuity conditions of the tangential field components $H_x$ and $E_y$ at each layer interface, a global system of linear equations is formed:
\begin{equation}
	\hat{\bf M}{\vec A}={\vec R},
	\label{eq:linsys}
\end{equation}
where $\hat{\bf M}$ is the matrix of the system of equations obtained from the boundary conditions, $\mathbf{A}$ is the vector of unknown amplitudes of waves and $\mathbf{R}$ is determined by the incident field. The amplitudes ${\vec A}$ finally enter the reflected field expansion
\begin{equation}
	\left[\begin{matrix}
		E_y^{(r)}\\ H_x^{(r)}
	\end{matrix}\right]=\!\!\!\sum_{m=-N}^{N} \left[\begin{matrix}
		1\\ k_{z;m}
	\end{matrix}\right] A_m^{(r)}\psi_m
	\exp(-\mathrm{i} k_{z;m}z),
	\label{eq:refl}
\end{equation}
where $k_{z;m}=(k_0^2-\kappa_x^2 m^2)^{1/2}$ (the branch $\mathrm{Im}\, k_{z;m}~\le~0$ is taken). It is evident that the map ${\vec\theta}\mapsto{\vec E}^{(r)}$ is completely determined by a sequence of algebraic operations and is therefore differentiable with respect to the design variables ${\vec\theta}$.

In the present work ${\vec\theta}$ denotes either the values of $\eps(x)$ on the computational grid (pixel-wise density parametrization) or the Fourier coefficients $\{\eps_m\}_{m=-N}^{N}$ of the permittivity (Fourier-parametrized projection). The gradient of the loss with respect to ${\vec\theta}$ is evaluated through the chain rule
\begin{equation}
	\frac{\partial\mathcal{L}}{\partial{\vec\theta}}=
	\frac{\partial\mathcal{L}}{\partial{\vec E}^{(r)}}\frac{\partial{\vec E}^{(r)}}{\partial\eps}\frac{\partial\eps}{\partial{\vec\theta}},
	\label{eq:chain}
\end{equation}
where every factor is computed by reverse-mode automatic differentiation. The mask parameters are updated by the standard gradient descent step
\begin{equation}
	{\vec\theta}^{(n+1)}={\vec\theta}^{(n)}-\eta\nabla_{\vec\theta}
	\mathcal{L}\bigl({\vec\theta}^{(n)}\bigr),
	\label{eq:gd}
\end{equation}
with learning rate $\eta$ and iteration counter $n$. The finite-difference approximation of the gradient requires $\mathcal{O}(N)$ forward evaluations per step, which is computationally prohibitive. Automatic differentiation reduces this cost to a single backward evaluation and makes optimization less resource-intensive.

\subsection{Pixel-Wise Density and Fourier-Parameterized Mask Reparametrizations}

The dielectric permittivity of the absorber layer of the mask is written as
\begin{align}
	\eps(x) = \eps_{\rm v} + \rho(x) (\eps_{\rm ab} - \eps_{\rm v}),
\end{align}
where $\eps_{\rm v}$ and $\eps_{\rm ab}$ are the dielectric permittivities of the vacuum and absorbing material, respectively, $\rho(x)$ is a function of the distribution of the absorber material in space.

Two complementary reparameterizations of the physical density $\rho(x)$ are investigated. In the pixel-wise density method an unconstrained auxiliary field $s(x)\in\mathbb{R}$ is optimized and mapped to the physical density through the sigmoid
\begin{equation}
	\rho(x)=\bigl(1+\exp[-s(x)]\bigr)^{-1}\in(0,1).
	\label{eq:sigmoid}
\end{equation}
The computational domain is divided into $N_p$ identical intervals (pixels) and the parameter $s(x_n)$ is optimized for each pixel $n$ with the coordinate $x_n$. The total loss
\begin{equation}
	\mathcal{L}= \lambda_{\rm d}\mathcal{L}_{\rm d}+ \lambda_{\rm bin}\mathcal{L}_{\rm bin}+ \lambda_{\rm tv}\mathcal{L}_{\rm tv}
	\label{eq:loss}
\end{equation}
is a combination of several components: a desired field matching term in the prescribed screen region ($\mathcal{L}_{\rm d} = \left(\left|{\vec E}^{(r)}\right|^2 - \left|{\vec E}^{(d)}\right|^2\right)^2$), a binarization penalty ($\mathcal{L}_{\rm bin} = \rho(1 - \rho)$) that drives the density towards the vertices of the interval [0, 1], and a total variation penalty ($\mathcal{L}_{\rm tv} = \sum_n (\rho(x_n) - \rho(x_{n+1}))^2$) that suppresses subwavelength artifacts. $\lambda_{\rm d}$, $\lambda_{\rm bin}$ and $\lambda_{\rm tv}$ are the weights of the loss terms. After convergence the continuous density is thresholded: $\rho_{\text{bin}}(x) = \{1 \text{ if } \rho(x) > 0.5; 0 \text{ if } \rho(x) \le 0.5\}$.

For the Fourier-parametrized projection the dense pixel representation is replaced by a band-limited latent function $f_{\rm latent}(x)=\mathrm{Re}\sum_{m=-N}^{N} a_m \psi_m$ optimized through the Fourier coefficients $a_m$, and is mapped to the density by a scaled sigmoid $\rho(x)=\bigl(1+\exp[-\beta f_{\rm latent}(x)]\bigr)^{-1}$ with steepness parameter $\beta$. The loss function retains the same physical, desired field, and binarization terms as in (\ref{eq:loss}); the total-variation penalty is replaced by a spectral penalty $\mathcal{L}_{\rm spec} = \sum_m m^2|a_m|^2$ on the high-order Fourier coefficients. Note that the band-limited nature of the latent function naturally suppresses high-frequency noise and forces the resulting binary mask to have smooth, well-defined walls.

\subsection{Waveguide Neural Operator}

The waveguide neural operator~(WGNO) replaces the solution of the linear system (\ref{eq:linsys}) by a multilayer perceptron ${\vec A}_{\vec\theta}$ that takes the right-hand side ${\vec R}$ and the Fourier coefficients of the permittivities $\{\eps_m^{(j)}\}_{j=1}^{J}$ as inputs and outputs an approximation of the amplitude vector ${\vec A}$~(see details in \cite{Eskin2025}). The WGNO is trained by minimizing the residual $\mathcal{L}_{\rm ph} = \|{\hat{\bf M}}{\vec A}_{\vec\theta}-{\vec R}\|_2^2$, so that the output of the neural network is forced to satisfy the physical laws. Therefore, the loss function (\ref{eq:loss}) must be expanded by a term $\mathcal{L}_{\rm ph}$. The optimization problem for ILT then takes the hybrid form
\begin{equation}
	\{{\vec\theta}^{\dagger},{\vec\theta}_{\rm NO}^{\dagger}\}
	= {\arg}\min_{\{{\vec\theta},{\vec\theta}_{\rm NO}\}}\,
	\left(\lambda_{\rm ph}\mathcal{L}_{\rm ph} + \mathcal{L}\right),
	\label{eq:wgno_opt}
\end{equation}
where ${\vec\theta}$ denotes the mask parameters and ${\vec\theta}_{\rm NO}$ the network weights. Importantly, the WGNO is mesh-independent, operates in the latent Fourier space, and inherits the physics-based structure of the waveguide method while addressing the linear algebraic bottleneck of the waveguide method.

\section{Numerical Experiments}

Numerical experiments were carried out for a realistic 2D EUV mask operating at the wavelength $\lambda{=}11.2$~nm. The mirror consisted of 30 Ru/Be/Sr periods (90 layers in total) with complex permittivities $\eps_{\rm Ru}=0.872-\mathrm{i}0.012$, $\eps_{\rm Be}=1.025-\mathrm{i}0.003$ and $\eps_{\rm Sr}=0.986-\mathrm{i}0.002$ for Ru, Be and Sr, respectively, at $\lambda{=}11.2$~nm. The permittivities of all media were obtained from the experimental database~\cite{HENKE1993181,CenterXRayOpt}. The thicknesses of these layers are 1.7 nm, 2.7 nm and 1.34 nm for Ru, Be and Sr, respectively. The angle of incidence was fixed at $6^\circ$ and $L_x = 214$ nm. Three absorber materials were compared in the inverse-design studies, namely TaBN ($\eps=0.909-\mathrm{i}0.060$), lanthanum ($\eps=1.090-\mathrm{i}0.033$) and uranium ($\eps=1.112-\mathrm{i}0.071$), all of which are representative of candidates under active investigation for EUV photomasks. The thickness of the absorber layer is 60 nm.

The training was performed on a single CPU node (Intel Core i5-9300H), optimization problems were implemented in PyTorch using the Adam optimizer with learning rate $\eta=10^{-3}$~\cite{kingma2014adam}. For WGNO, we used an MLP with 2 hidden layers and 128 neurons each with hyperbolic tangent activation function ($\tanh$).

In our experiments, the desired intensity of the electric field $I^{(d)}$ is taken to be binary, equal to one in some regions and zero in others. We took the dependence of the electric field on the $x$-coordinate in the target plane, $S$, to be the same as that of the incident field, $E_y^{(d)}=\exp(-{\rm i}k_{0x}x)$. This desired field ensured phase consistency of the field between the individual sections of the target plane and significantly improved the optimization process. Next, we decomposed this field, $E_y^{(d)}$, into free-space waves, taking into account its periodicity in the $x$ direction. In this expansion all non-propagating components were removed, and the resulting propagating components formed a new intensity pattern $\tilde{I}^{(d)}$, which was then used to calculate the loss~(\ref{eq:loss}).

\subsection{Pixel-Wise Density Optimization}

The pixel-wise density optimization starts from an unconstrained auxiliary field $s(x)$ and converges to a continuous physical density $\rho(x)$ that, after hard binarization at threshold $0.5$, produces a physically admissible mask profile. The number of pixels is $N_p=500$. For this method of optimization weights of the loss terms are taken as $\lambda_{\rm d} = 10^6$, $\lambda_{\rm bin} = 1$ and $\lambda_{\rm tv} = 10^{-3\left(1+ n/N_{\rm ep}\right)}$, where $n$ is the current epoch of optimization and $N_{\rm ep}$ is the total number of epochs.

For the TaBN absorber, the central peak intensity on the screen grows after profile binarization and the physically manufactured binary structure preserves the desired optical effect. As the number of optimization epochs increases, the mask absorber density converges to the binarized density, while the intensity becomes more focused and the side lobes are visibly suppressed. The pixel-wise density method serves as a baseline for the more compact Fourier-parametrized strategy introduced next.

\begin{figure}[t]
	\centering
	\includegraphics[width=\columnwidth]{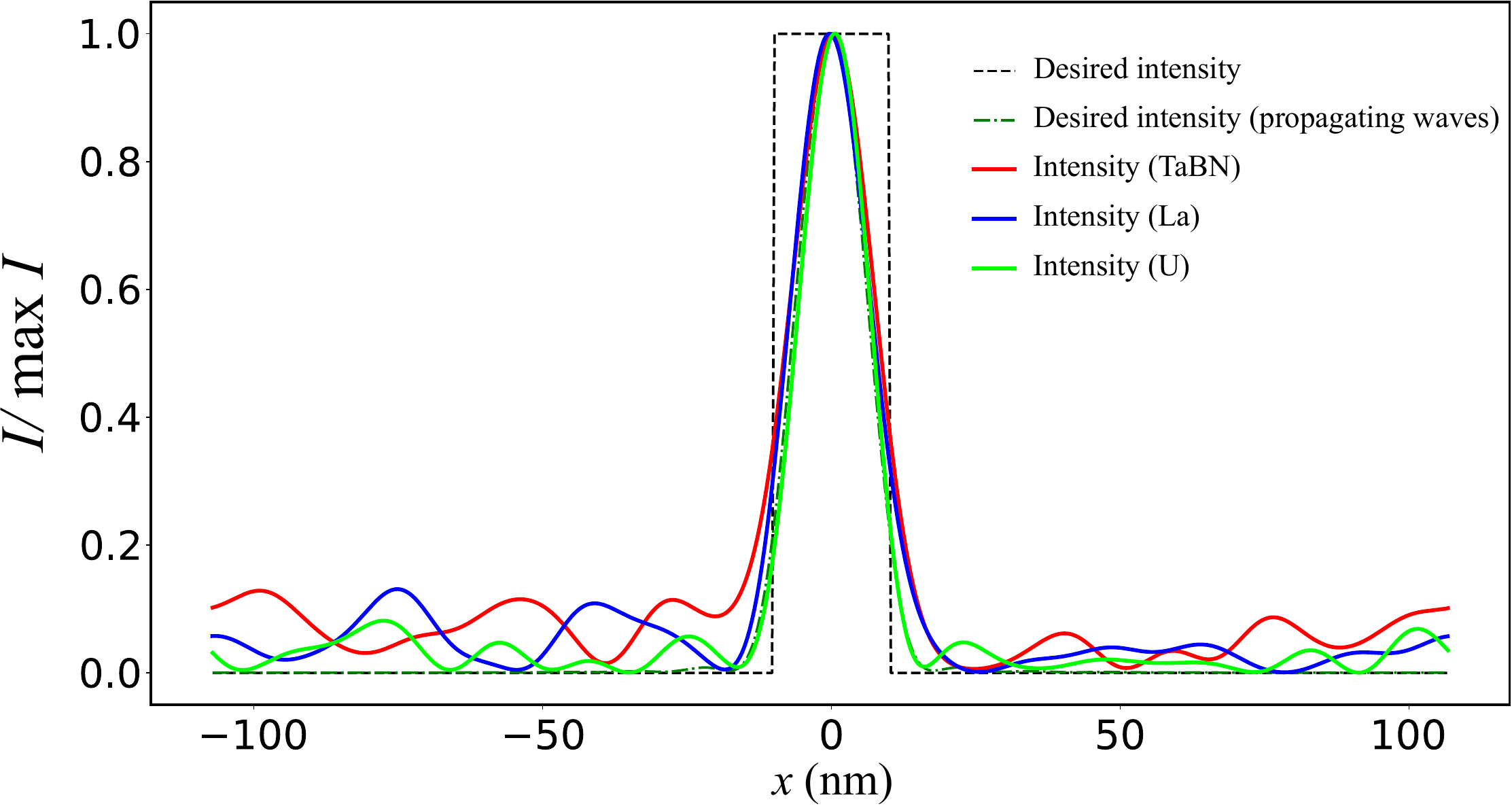}
	\caption{Electric field intensity on the surface $S$ normalized to the maximum value of the field: the desired field, the desired field considering only propagating waves, and the scattered field for TaBN, La, and U absorber materials. The ratios of the maximum field intensity to the incident field intensity ($\max I / I_0$) are 0.64, 1.99, and 0.45 for TaBN, La, and U, respectively.}
	\label{fig:absorbers}
\end{figure}

The choice of absorber material has a direct impact on the resulting image quality. The intensity distributions on plane $S$ obtained for TaBN, lanthanum and uranium at $N{=}8$ harmonics and $5000$ optimization epochs share the same generic peak topology imposed by the target, but the relative height of the central maximum and the structure of the side lobes differ noticeably between the three materials (see Fig.~\ref{fig:absorbers}). Lanthanum provides the best central maximum, whereas uranium yields the closest match to the desired field.

\begin{figure}[t]
	\centering
	\includegraphics[width=\columnwidth]{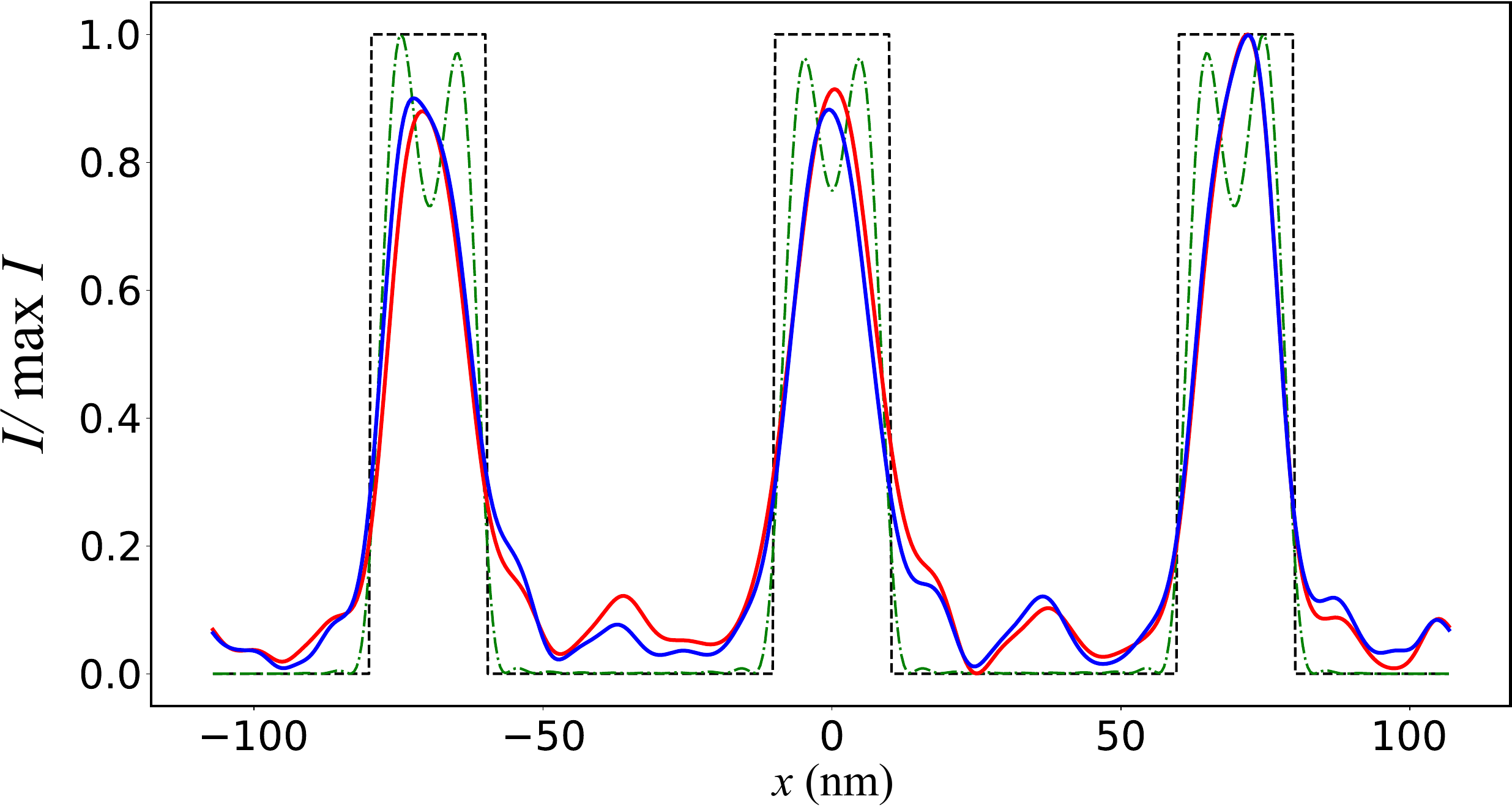}
	\caption{The electric field intensity on the surface $S$ normalized to the maximum field value: the black dashed line is the desired field, the green dashed-dotted line is the desired field taking into account only propagating waves, the red solid line is the scattered field obtained using the pixel-wise optimization method, and the blue solid line is the field obtained using the Fourier-parameterized optimization method.}
	\label{fig3}
\end{figure}

The pixel-wise density method generalizes in a straightforward manner to multi-strip targets. For a three-strip target, the lanthanum absorber recovers three distinct intensity maxima on the surface $S$ whose positions match the prescribed geometry, as shown in Fig.~\ref{fig3} for $N=25$. This configuration
mimics the dense parallel interconnects encountered in real backside power delivery networks and clock distribution lines, and it demonstrates that the present ILT pipeline scales to the multi-feature layouts of practical interest.

\subsection{Fourier-Parameterized Optimization}

The Fourier-parametrized projection with $N{=}25$ harmonics recovers the same three-peak target as the pixel-wise density method with $N{=}25$ (see Fig.~\ref{fig3}), but with one important quantitative advantage. The wall-clock optimization time decreases from $179$~seconds (pixel-wise) to $137$~seconds (Fourier-parameterized), i.e.\ a $1.31$-fold acceleration, despite the fact that the two methods solve exactly the same forward problem. The improvement is attributed to the more compact optimization space, to the suppression of high-frequency artifacts, and to the smoothness of the resulting binary mask, which exhibits well-defined walls without jagged pixelated boundaries. 

Note that replacing the forward solver with the WGNO gives comparable optimization results. However, the WGNO currently does not provide a speed advantage, as the neural network is trained simultaneously with the mask optimization. This means that the computational cost of training outweighs the benefit of the faster forward solve.

\subsection{3D Mask Optimization}

\begin{figure}[t]
	\centering
	\includegraphics[width=\columnwidth]{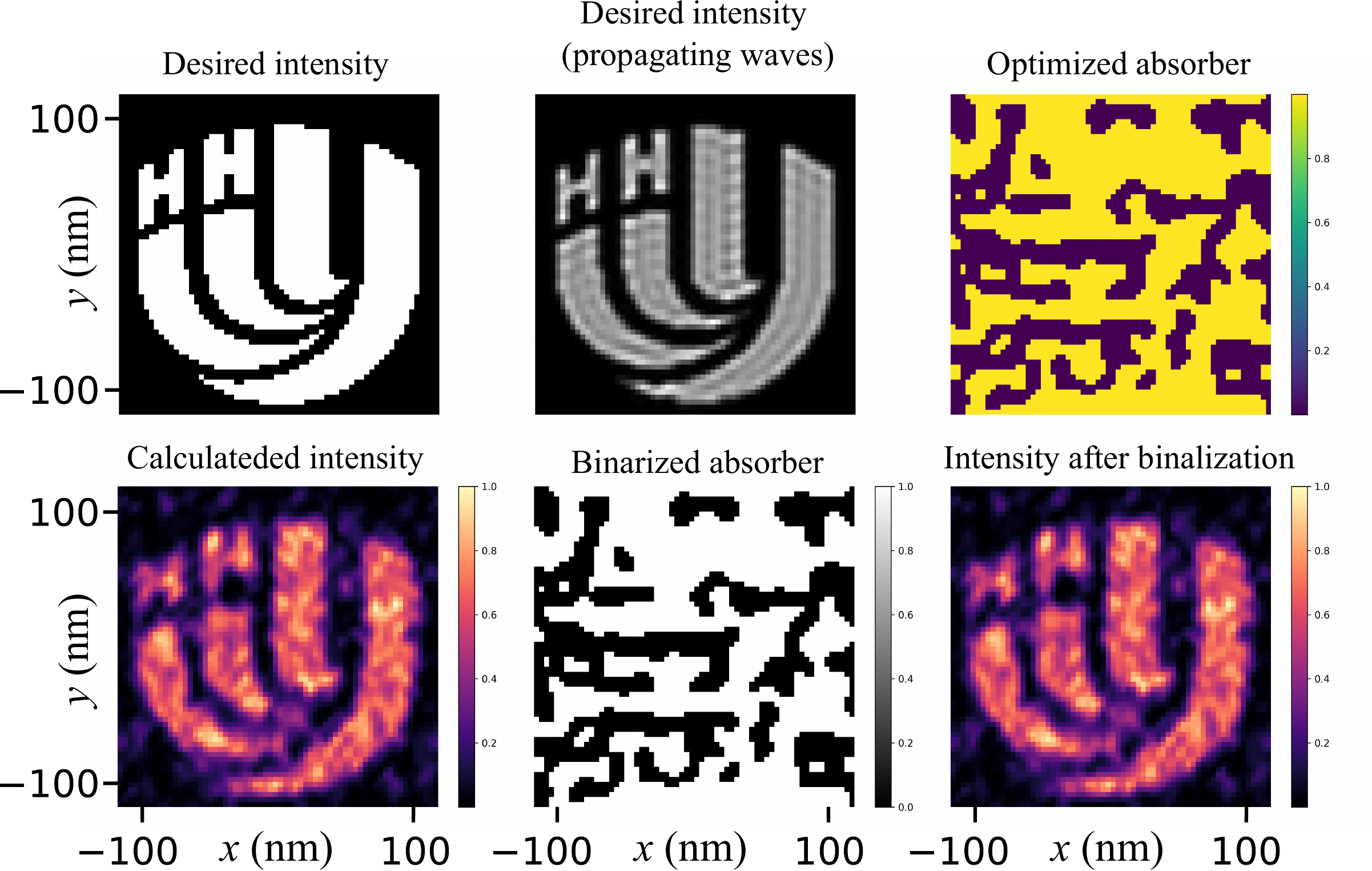}
	\caption{3D Mask Optimization for the desired pattern of size $214{\times}214$~nm and absorber TaBN.}
	\label{fig4}
\end{figure}

The 2D framework extends naturally to three dimensions. The forward simulation is performed with the three-dimensional waveguide method or the WGNO, and the inverse problem is solved by the same density and
Fourier-parametrized strategies as in the 2D case. In the 3D formulation the one-dimensional harmonic index $N$ is replaced by a two-dimensional grid $N_x{\times}N_y$. For a target pattern of size $214{\times}214$~nm with $N_x{=}N_y{=}32$ and a TaBN absorber, the method reproduces the prescribed intensity distribution on the screen after 500 epochs of optimization, and the Fourier-parametrized version produces the most regular and manufacturable mask profile (see Fig.~\ref{fig4}). These experiments confirm that the proposed ILT framework inherits the scalability of the waveguide method and the WGNO and is applicable to realistic EUV mask design.

\section{Conclusion}

In this work, we have presented a gradient-based inverse lithography framework for EUV masks in which the differentiable waveguide method and the waveguide neural operator serve as end-to-end physics engines. The permittivity of the absorber is recovered by automatic differentiation of the full forward diffraction model, using either a pixel-wise density reparameterization or a Fourier-parametrized projection, both followed by hard binarization. Numerical experiments on realistic 2D and 3D masks at $\lambda{=}11.2$~nm with three absorber
materials (TaBN, La, U) demonstrate that both strategies recover mask structures producing the desired intensity distribution on the wafer. The Fourier-parametrized strategy achieves a $1.31$-fold speed-up over the pixel-wise method ($137$~s vs.\ $179$~s) and yields smoother, more manufacturable mask walls, while the comparison of absorber materials reveals that lanthanum provides the best central maximum and uranium yields the closest match to the desired field. Preliminary 3D optimizations at $N_x{\times}N_y{=}32{\times}32$ confirm the scalability of the
proposed pipeline. These results establish a fully differentiable route to EUV inverse mask design in which the rigorous forward model is preserved throughout the optimization process, and indicate that the same framework is applicable to the inverse design of metamaterials and other periodic electromagnetic structures.


\begin{thebibliography}{9}
	
	\bibitem{it7} N.\,I.\;Chkhalo, New concept for the development of high-performance X-ray lithography, \textit{Russ. Microelectron.}, \textbf{53}, 5, 397--407 (2024).
	
	\bibitem{Pang2021} L.\;Pang, Inverse lithography technology: 30 years from concept to practical, full-chip reality, \textit{J. Micro/Nanopatterning, Mater. Metrol.}, \textbf{20}, 3, 030901 (2021).
	
	\bibitem{it1} H.\;Tanabe, A.\;Jinguji, and A.\;Takahashi, Accelerating extreme ultraviolet lithography simulation with weakly guiding approximation and source position dependent transmission cross coefficient formula, \textit{J. Micro/Nanopatterning, Mater. Metrol.}, \textbf{23}, 1, 14201 (2024).
	
	\bibitem{raissi2019physics} M.\;Raissi, P.\;Perdikaris, and G.\,E.\;Karniadakis, Physics-informed neural networks: a deep learning framework for solving forward and inverse problems involving nonlinear partial differential equations, \textit{J. Comput. Phys.}, \textbf{378}, 686--707, (2019).
	
	\bibitem{eskin2024causal} V.\,A.\;Es'kin, D.\,V.\;Davydov, E.\,D.\;Egorova, A.\,O.\;Malkhanov, M.\,A.\;Akhukov, and M.\,E.\;Smorkalov, About modifications of the loss function for the causal training of physics-informed neural networks, \textit{Dokl. Math.}, \textbf{110}, S1, S172--S192 (2024).
	
	\bibitem{medvedev2024} V.\;Medvedev, A.\;Erdmann, and A.\;Rosskopf, 3D EUV mask simulator based on physics-informed neural networks: effects of polarization and illumination, \textit{Comput. Opt.}, \textbf{13023}, 19--36 (2024).
	
	\bibitem{Eskin2025} V.\,A.\;Es'kin and E.\,V.\;Ivanov, Physics-informed neural networks and neural operators for a study of EUV electromagnetic wave diffraction from a lithography mask, in \textit{2025 Days on Diffraction}, pp. 48--53, (2025).
	
	
	\bibitem{lucas1996efficient} K.\,D.\ Lucas, H.\ Tanabe, and A.\,J.\ Strojwas, Efficient and rigorous three-dimensional model for optical lithography simulation, \textit{J. Opt. Soc. Am. A}, \textbf{13}, 11, 2187--2199 (1996).
	
	\bibitem{kingma2014adam} D.\,P.\ Kingma and J.\ Ba, Adam: a method for stochastic optimization, in \textit{Proceedings of the 3rd International Conference on Learning Representations (ICLR)}, (2015).
	
	\bibitem{HENKE1993181} B.\,L.\ Henke, E.\,M.\ Gullikson, and J.\,C.\ Davis, X-ray interactions: photoabsorption, scattering, transmission, and reflection at E = 50--30000 eV, Z = 1--92, \textit{Atomic Data and Nuclear Data Tables}, \textbf{54}, 2, 181--342 (1993).
	
	\bibitem{CenterXRayOpt} {Center for X-Ray Optics. Lawrence Berkeley National Laboratory}. (1993--2025) Index of refraction. \url{https://henke.lbl.gov/optical_constants/getdb2.html}
	
\end{thebibliography}
\end{document}